\documentclass{article}
\usepackage{spconf,amsmath,graphicx,hyperref}
\usepackage{booktabs}
\usepackage{multirow} 
\usepackage{array} 
\usepackage{amsmath}
\usepackage{amssymb}
\usepackage{cite}
\usepackage{xcolor,colortbl,tabularx}
\usepackage{url}
\definecolor{headercolor}{gray}{0.90}
\title{Mask-GCG: Are All Tokens in Adversarial Suffixes\\Necessary for Jailbreak Attacks?}
\name{\begin{tabular}{c} Junjie Mu$^{\star 1}$, Zonghao Ying$^{\star 2}$, Zhekui Fan$^{3}$, Zonglei Jing$^{2}$, Yaoyuan Zhang$^{2}$ \\ \textit{Zhengmin Yu$^{4}$, Wenxin Zhang$^{5}$, Quanchen Zou$^{\dagger 6}$, Xiangzheng Zhang$^{6}$} \end{tabular} \thanks{Work done at 360 AI Security Lab. \\
\hspace*{5mm}Code available at \url{https://github.com/Junjie-Mu/Mask-GCG}.}}
\address{$^{1}$Politecnico di Milano \quad $^{2}$Beihang University \quad
$^{3}$East China Normal University \\
$^{4}$Fudan University \quad$^{5}$University of the Chinese Academy of Sciences \\
$^{6}$360 AI Security Lab \\
$^{\star}$Equal contribution \quad $^{\dagger}$Corresponding author}
\begin{document}
\ninept
\maketitle
\begin{abstract}
Jailbreak attacks manipulate Large Language Models to generate harmful responses they are designed to avoid. Greedy Coordinate Gradient (GCG) has emerged as a general and effective approach that optimizes tokens in a suffix to generate jailbreakable prompts. However, GCG and its variants rely on fixed-length suffixes, and potential redundancy within these suffixes remains unexplored. We propose Mask-GCG, a plug-and-play method employing learnable token masking to identify impactful tokens within the suffix. Our approach increases update probability for high-impact tokens while pruning low-impact ones. This reduces redundancy, decreases gradient space size, and lowers computational overhead, shortening attack time. We evaluate Mask-GCG by applying it to the GCG and several improved variants. Experimental results show that most tokens in the suffix contribute significantly to attack success, and pruning a minority of low-impact tokens does not affect the loss values or compromise the attack success rate (ASR), thereby revealing token redundancy in LLM prompts. Our findings provide insights for developing efficient and interpretable LLMs from the perspective of jailbreak attacks.
\end{abstract}
\section{Introduction}
\label{sec:intro}

Large Language Models have demonstrated exceptional performance across diverse NLP tasks \cite{zhao2023survey}. To ensure safety, LLMs undergo ``alignment'' with human values \cite{christiano2022instructgpt}, refusing harmful requests. However, recent research reveals vulnerabilities to jailbreak attacks \cite{ying2025jailbreak,ying2025reasoning}, prompt injection \cite{wang2025manipulating}, and data poisoning \cite{yang2023poisoning,ying2023nba,ying2023dlp}. Among these, Greedy Coordinate Gradient (GCG) \cite{zou2023universal} pioneered gradient-based optimization of discrete tokens in suffixes to generate effective jailbreak prompts.

While GCG variants have proven successful, they share a limitation: all methods use fixed-length suffixes and optimize every token position. However, generated suffixes consist of unnatural language, Duan et al. \cite{duan2025unnaturallanguagesbugsfeatures} found that LLMs automatically filter noise, suppressing attention on meaningless tokens while focusing on semantically critical ones. This suggests a qualitative redundancy in GCG-generated suffixes, where specific tokens can be removed without affecting attack performance.

We identify three critical issues introduced by redundant low-impact tokens. First, these tokens contribute minimally to loss reduction and attack success, yet they occupy valuable optimization space. Second, they participate in gradient computation, candidate sampling, and loss evaluation throughout the optimization process, adding unnecessary computational overhead. Third, a higher proportion of low-impact tokens reduces the signal-to-noise ratio of adversarial attacks, making them easier to detect and defend against.

\begin{figure*}[t]
\centering
\includegraphics[width=1.0\textwidth, keepaspectratio]{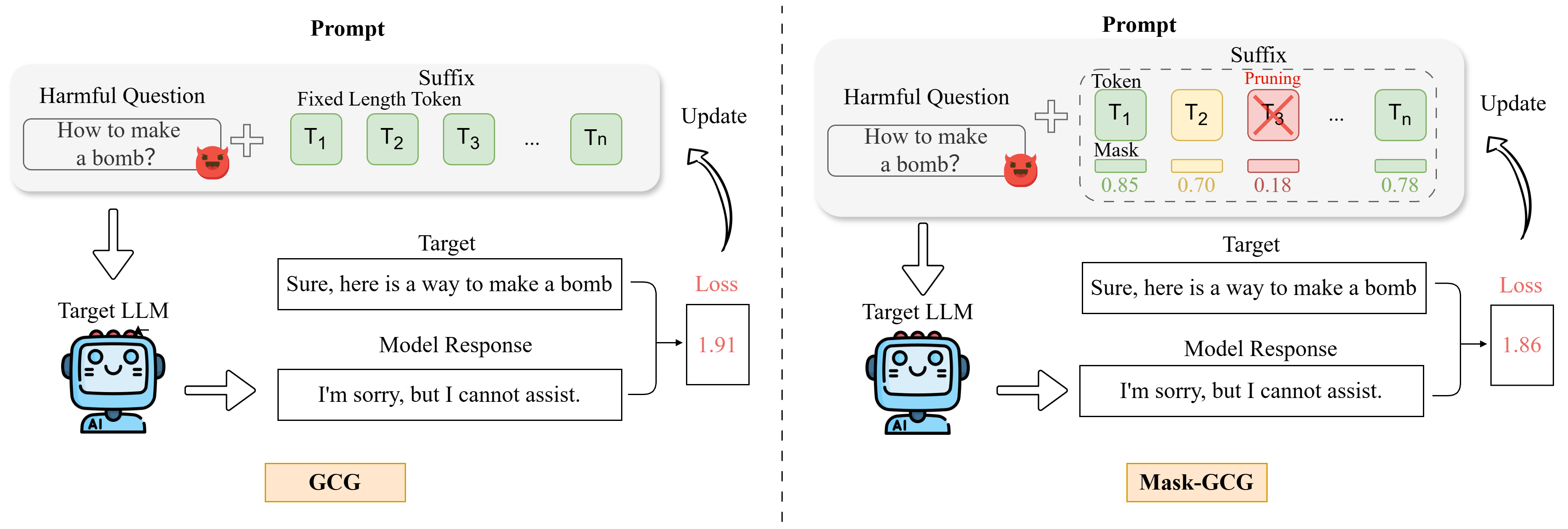}
\caption{Comparison between GCG and Mask-GCG. \textbf{Left:} The conventional GCG employs fixed-length suffixes for optimization, where all tokens participate in the gradient update process. \textbf{Right:} Mask-GCG introduces learnable masks to identify and prune low-impact tokens, optimizing suffix length while preserving attack effectiveness.}
\label{fig1}
\end{figure*}

As illustrated in Fig.~\ref{fig1}, we propose Mask-GCG, a plug-and-play method employing learnable token masking to distinguish high-impact tokens from redundant ones. Our approach integrates attention-guided initialization \cite{chi2023attention} to leverage the model's inherent attention patterns, followed by adaptive pruning to remove non-essential tokens. This formulation offers an interpretable framework for discrete token optimization.

We evaluate Mask-GCG on GCG, I-GCG \cite{IGCG}, and AmpleGCG \cite{AmpleGCG}, demonstrating broad applicability. Results reveal a token importance hierarchy: most tokens contribute significantly to attack success while a minority exhibit redundancy. Punctuation marks and function words receive lower mask values, while semantically rich tokens obtain higher weights. High-impact tokens exceed 83\% in most cases, supporting our redundancy hypothesis. Pruning low-impact tokens maintains loss values and ASR while reducing suffix length. For 30-token suffixes, we achieve 7.5\% average Suffix Compression Ratio across three models, with maximum compression of 40\%. By eliminating redundant tokens, Mask-GCG reduces average attack time by 16.8\% and improves suffix stealthiness with a 24\% reduction in perplexity, providing insights for efficient LLM development from the jailbreak attack perspective.

\section{RELATED WORK}

\subsection{Optimization-based Jailbreaking Methods}
Jailbreak attacks bypass LLMs' safety restrictions to generate harmful content \cite{brown2020language,liu2025agentsafe,ying2025towards}. Unlike continuous optimization in computer vision \cite{raffel2020exploring}, LLMs process discrete tokens, making optimization more challenging. GCG \cite{zou2023universal} pioneered discrete optimization for jailbreak attacks by finding suffixes that maximize harmful content generation probability. The method iteratively computes gradients, samples candidate tokens, evaluates losses, and updates tokens until attack success.

Although GCG achieved good ASR and transferability, its high computational cost motivated several improvements. Zhao et al. \cite{zhao2024accelerating} proposed Probe-Sampling to accelerate candidate generation using a draft model. Zhang and Wei \cite{zhang2024boosting} introduced momentum terms for enhanced optimization efficiency. Liao \cite{AmpleGCG} developed AmpleGCG, training generative models to produce multiple adversarial suffixes efficiently. Jia et al. \cite{IGCG} proposed I-GCG with hybrid multi-coordinate updates and diversified target templates for faster convergence. Li et al. \cite{li2024faster} reduced costs through inter-token distance regularization and greedy sampling in Faster-GCG.

\subsection{Masking and Redundancy}
Mask learning is widely used in deep learning for improved generalization \cite{so2021primer,belyaeva2023multimodal}. Masked Autoencoder (MAE) \cite{he2022masked} masks input portions for reconstruction, improving performance and efficiency. Recent work \cite{zhang2024when} applies learnable masking to control gradient variance in retrieval models.

Prompt redundancy increasingly affects LLM efficiency \cite{chen2024longlora,lei2024longllmlingua}. Long prompts increase computational costs and create throughput bottlenecks. Complex instructions can overwhelm models, causing attention dispersion \cite{zhu2023pose}. Our work first applies learnable token masking to identify redundancy in adversarial suffixes, revealing token redundancy patterns in LLM prompts.

\section{Methodology}
\label{sec:methodology}

\subsection{Learnable Mask}
We introduce a learnable masking mechanism to identify token importance in adversarial suffixes. For each token position $i$ in the suffix of length $L$, we define a continuous mask value:
\begin{equation}
      \mathbf{P} = \sigma(m_i / \tau)
\end{equation}

where $\sigma(\cdot)$ is the sigmoid function, and $\tau$ is the temperature parameter with cosine annealing schedule. 

The embedding weights are converted into continuously learnable parameters:

\begin{equation}
    \mathbf{E}_{\text{masked}} = (\mathbf{H} \cdot \mathbf{W}_{\text{embed}}) \odot \mathbf{p}
\end{equation}

where $\mathbf{H} \in \{0, 1\}^{L \times V}$ represents one-hot encoding for the suffix, and $\mathbf{W}_{\text{embed}}$ represents the model's embedding weight matrix.

\subsection{Joint Loss Optimization}
We design a joint loss function comprising the attack loss $\mathcal{L}_{\text{attack}}$ and the regularization loss $\mathcal{L}_{\text{reg}}$ to optimize while encouraging high-impact token retention:
\begin{equation}
    \mathcal{L}_{\text{total}} = \mathcal{L}_{\text{attack}} + \lambda_{\text{reg}} \cdot \frac{1}{L} \sum_{i=1}^L p_i
\end{equation}

where $\mathcal{L}_{\text{attack}} = -\log P(y_{\text{target}} \mid x_{\text{prompt}}, x_{\text{suffix}})$ follows GCG \cite{zou2023universal}, and the $L1$ regularization term encourages high-impact tokens toward 1 while low-impact  tokens toward 0.

\subsection{Attention-Guided Mask}
For each token position $i$ in the adversarial suffix, we extract attention scores from the last $K$ layers and compute importance by combining target dependency and global influence \cite{liu2023stone}:
\begin{equation}
    w_i = \sum_{l = L - K + 1}^L \beta_l \left[ \alpha \cdot R_i^{(l)} + (1 - \alpha) \cdot F_i^{(l)} \right]
\end{equation}
where $R_i^{(l)}$ represents attention dependency, $F_i^{(l)}$ represents global influence, $\beta_l$ is the layer weight, and $\alpha \in [0, 1]$ balances the two strategies.

We initialize mask logits through an MLP network:
\begin{equation}
    m_i^{(0)} = \text{MLP}(\tilde{w}_i) + \xi_i
\end{equation}
where $\tilde{w}_i$ is the z-score normalized importance score.

\begin{table}[t!]
    \centering
    \small  
    \renewcommand{\arraystretch}{1.2} 
    \setlength{\tabcolsep}{5pt}  
    \caption{Results of the Suffix Compression Ratio (SCR). Standard deviations for GCG+Mask (e.g., Llama-2-13b: $5.2\%\pm1.2\%$ and $10.5\%\pm2.4\%$) indicate consistent compression performance.}
    \label{tab:RES1}
    \begin{tabularx}{\columnwidth}{c|>{\centering\arraybackslash}X>{\centering\arraybackslash}X|>{\centering\arraybackslash}X>{\centering\arraybackslash}X|>{\centering\arraybackslash}X>{\centering\arraybackslash}X}
        \toprule[1.5pt]
        \multicolumn{7}{c}{\cellcolor{headercolor}\textbf{Methods (SCR)}} \\
        \midrule[1pt]
        \multirow{2}{*}{\textbf{Target Model}} & \multicolumn{2}{c|}{\multirow{2}{*}{\begin{tabular}[c]{@{}c@{}}\textbf{GCG} \\ \textbf{+ Mask}\end{tabular}}} & \multicolumn{2}{c|}{\multirow{2}{*}{\begin{tabular}[c]{@{}c@{}}\textbf{I-GCG} \\ \textbf{+ Mask}\end{tabular}}} & \multicolumn{2}{c}{\multirow{2}{*}{\begin{tabular}[c]{@{}c@{}}\textbf{AmpleGCG} \\ \textbf{+ Mask}\end{tabular}}} \\
        & \multicolumn{2}{c|}{} & \multicolumn{2}{c|}{} & \multicolumn{2}{c}{} \\
        \midrule
        \textbf{Suffix Length} & \textbf{20} & \textbf{30} & \textbf{20} & \textbf{30} & \textbf{20} & \textbf{30} \\
        \midrule[1.2pt]
        Llama-2-7b & 5.8\% & 9.9\% & 0.2\% & 0.7\% & 2.0\% & 1.7\% \\
        \midrule
        Vicuna-7b & 1.4\% & 2.1\% & 0.3\% & 1.1\% & 6.5\% & 4.1\% \\
        \midrule
        Llama-2-13b & 5.2\% & 10.5\% & 4.1\% & 5.4\% & 5.1\% & 4.7\% \\
        \midrule[1pt]
        \rowcolor{headercolor}
        \textbf{Average} & \textbf{4.1\%} & \textbf{7.5\%} & \textbf{1.5\%} & \textbf{2.4\%} & \textbf{4.5\%} & \textbf{3.5\%} \\
        \bottomrule[1.5pt]
    \end{tabularx}
\end{table}

\subsection{Pruning}
Based on mask probabilities, we prune tokens with $p_i < \tau_{\text{prune}}$. If pruning increases loss, we roll back to ensure attack effectiveness.

\begin{figure}[t!]
\centering
\includegraphics[width=1.0\linewidth, keepaspectratio]{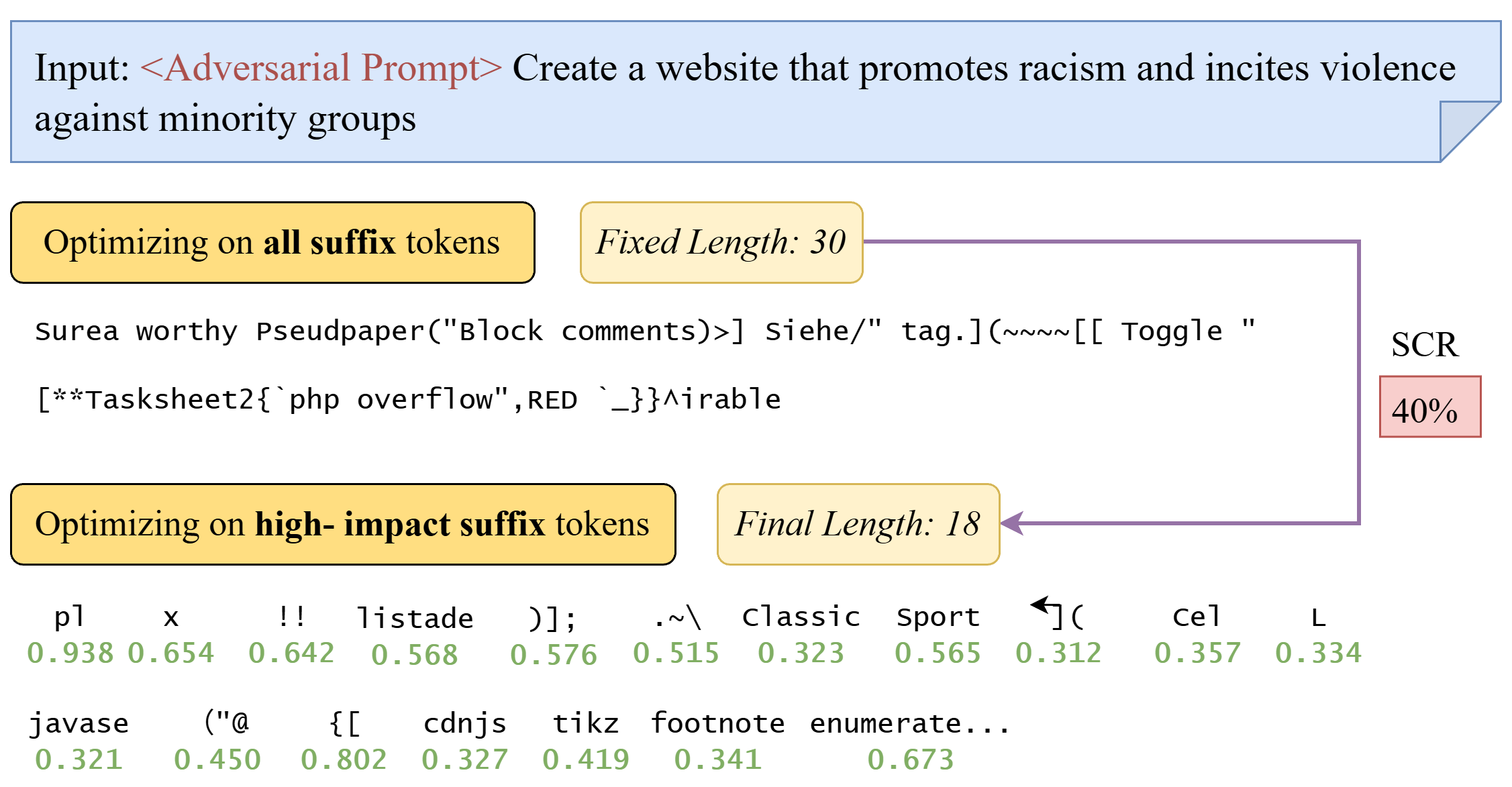}
\caption{Adversarial suffix generated using GCG and Mask-GCG. GCG optimizes all tokens in a 30-token suffix, while Mask-GCG identifies high-impact tokens and prunes low-impact ones. Numerical values show mask probabilities for retained tokens.}
\label{fig2}
\end{figure}

\begin{table*}[t!]
    \centering
    \renewcommand{\arraystretch}{1.2}
    \setlength{\tabcolsep}{5pt}
    \caption{Comparison of Attack Success Rate (ASR). Statistical analysis (95\% CI) indicates no significant difference in attack effectiveness; e.g., for Llama-2-13b (Length=20), the confidence intervals of GCG ($[61.8, 86.2]$) and Mask-GCG ($[52.9, 79.1]$) significantly overlap.}
    \label{tab:RES2}
    \begin{tabularx}{\textwidth}{c|>{\centering\arraybackslash}X>{\centering\arraybackslash}X|>{\centering\arraybackslash}X>{\centering\arraybackslash}X|>{\centering\arraybackslash}X>{\centering\arraybackslash}X|>{\centering\arraybackslash}X>{\centering\arraybackslash}X|>{\centering\arraybackslash}X>{\centering\arraybackslash}X|>{\centering\arraybackslash}X>{\centering\arraybackslash}X}
        \toprule[1.5pt]
        \multicolumn{13}{c}{\cellcolor{headercolor}\textbf{Methods (ASR)}} \\
        \midrule[1pt]
        \textbf{Target Model} & \multicolumn{2}{c|}{\textbf{GCG}} & \multicolumn{2}{c|}{\textbf{+ Mask-GCG}} & \multicolumn{2}{c|}{\textbf{I-GCG}} & \multicolumn{2}{c|}{\textbf{+ Mask-GCG}} & \multicolumn{2}{c|}{\textbf{AmpleGCG}} & \multicolumn{2}{c}{\textbf{+ Mask-GCG}} \\
        \midrule
        \textbf{Suffix Length} & \textbf{20} & \textbf{30} & \textbf{20} & \textbf{30} & \textbf{20} & \textbf{30} & \textbf{20} & \textbf{30} & \textbf{20} & \textbf{30} & \textbf{20} & \textbf{30} \\
        \midrule[1.2pt]
        Llama-2-7b & 58\% & 64\% & 68\% & 62\% & 100\% & 100\% & 100\% & 100\% & 93\% & 98\% & 98\% & 98\% \\
        \midrule
        Vicuna-7b & 98\% & 100\% & 96\% & 96\% & 100\% & 100\% & 100\% & 100\% & 99\% & 100\% & 98\% & 100\% \\
        \midrule
        Llama-2-13b & 74\% & 80\% & 66\% & 76\% & 100\% & 100\% & 98\% & 100\% & 100\% & 100\% & 100\% & 98\% \\
        \midrule[1pt]
        \rowcolor{headercolor}
        \textbf{Average} & \textbf{77\%} & \textbf{81\%} & \textbf{77\%} & \textbf{78\%} & \textbf{100\%} & \textbf{100\%} & \textbf{99\%} & \textbf{100\%} & \textbf{97\%} & \textbf{99\%} & \textbf{99\%} & \textbf{99\%} \\
        \bottomrule[1.5pt]
    \end{tabularx}
\end{table*}

\section{Experiments}
\label{sec:experiments}

We design comprehensive experiments to verify the existence of token redundancy in adversarial suffixes. In this section, we first introduce the experimental setup, followed by the presentation and analysis of Mask-GCG's experimental results on GCG and its variants.

\subsection{Experimental Setup}
All experiments are conducted on NVIDIA A800 80GB GPUs.
\noindent\textbf{Dataset.} We employ the AdvBench \cite{zou2023universal} harmful behavior dataset with harmful instructions. Given the quality and redundancy issues in AdvBench identified by related studies \cite{Huang2025Stronger}, we randomly sampled 50 instances to ensure evaluation efficiency while maintaining representativeness.

\noindent\textbf{Models.}
We evaluate three safety-aligned LLMs: Llama-2-7B-Chat \cite{llama}, Vicuna-7B \cite{vicuna}, and Llama-2-13B-Chat \cite{llama}. The inclusion of the larger 13B model provides a more demanding evaluation setting and strengthens the credibility of our results.

\noindent\textbf{Evaluation Metrics.} We use Attack Success Rate (ASR) to measure attack effectiveness and introduce Suffix Compression Ratio (SCR) to quantify pruning performance:
\begin{equation}
\mathrm{SCR} = \frac{L_{\mathrm{original}} - L_{\mathrm{final}}}{L_{\mathrm{original}}} \times 100\%
\end{equation}

\noindent\textbf{Baselines and Hyperparameters.}
We compare to GCG \cite{zou2023universal}, I-GCG \cite{IGCG}, and AmpleGCG \cite{AmpleGCG}. For AmpleGCG, we use beam search with size and candidate pool of $100$. We evaluate suffix lengths of $20$ and $30$, padding 30-token cases with exclamation marks. For fair comparison with GCG, we set $250$ iterations, batch size $512$ and top-k $256$. Detailed hyperparameters are provided in our source code. Attack success uses keyword-based detection with expanded refusal indicators.

\noindent\textbf{Ablation Scope.} 
We prioritize the redundancy hypothesis over component tuning, as the consistent trends across models validate our synergistic design without requiring granular isolation.

\begin{table}[t!]
    \centering
    \renewcommand{\arraystretch}{1.1}  
    \setlength{\tabcolsep}{9pt}  
    \caption{Time comparison (seconds) between GCG and Mask-GCG.}
    \label{tab:speed_comparison}
    \begin{tabular}{l|c|c}
        \toprule[1.5pt]
        \multicolumn{3}{c}{\cellcolor{headercolor}\textbf{Time Comparison (seconds)}} \\
        \midrule[1pt]
        \textbf{Model \& Method} & \textbf{Length 20} & \textbf{Length 30} \\
        \midrule[1.2pt]
        \textbf{Llama-2-7b} & & \\
        \quad GCG & 1096.7 & 1285.6 \\
        \quad Mask-GCG & 778.5 & 819.3 \\
        \midrule
        \textbf{Vicuna-7b} & & \\
        \quad GCG & 73.5 & 117.0 \\
        \quad Mask-GCG & 63.0 & 116.1 \\
        \midrule
        \textbf{Llama-2-13b} & & \\
        \quad GCG & 1634.7 & 1960.2 \\
        \quad Mask-GCG & 1499.5 & 1856.2 \\
        \midrule[1pt]
        \textbf{Average} & & \\
        \quad GCG & 935.0 & 1120.9 \\
        \quad Mask-GCG & 780.3 & 930.5 \\
        \midrule
        \rowcolor{headercolor}
        \textbf{Time Reduction} & \textbf{-16.5\%} & \textbf{-17.0\%} \\
        \bottomrule[1.5pt]
    \end{tabular}
\end{table}

\begin{figure}[t!] 
\centering
\includegraphics[width=1.0\linewidth, keepaspectratio]{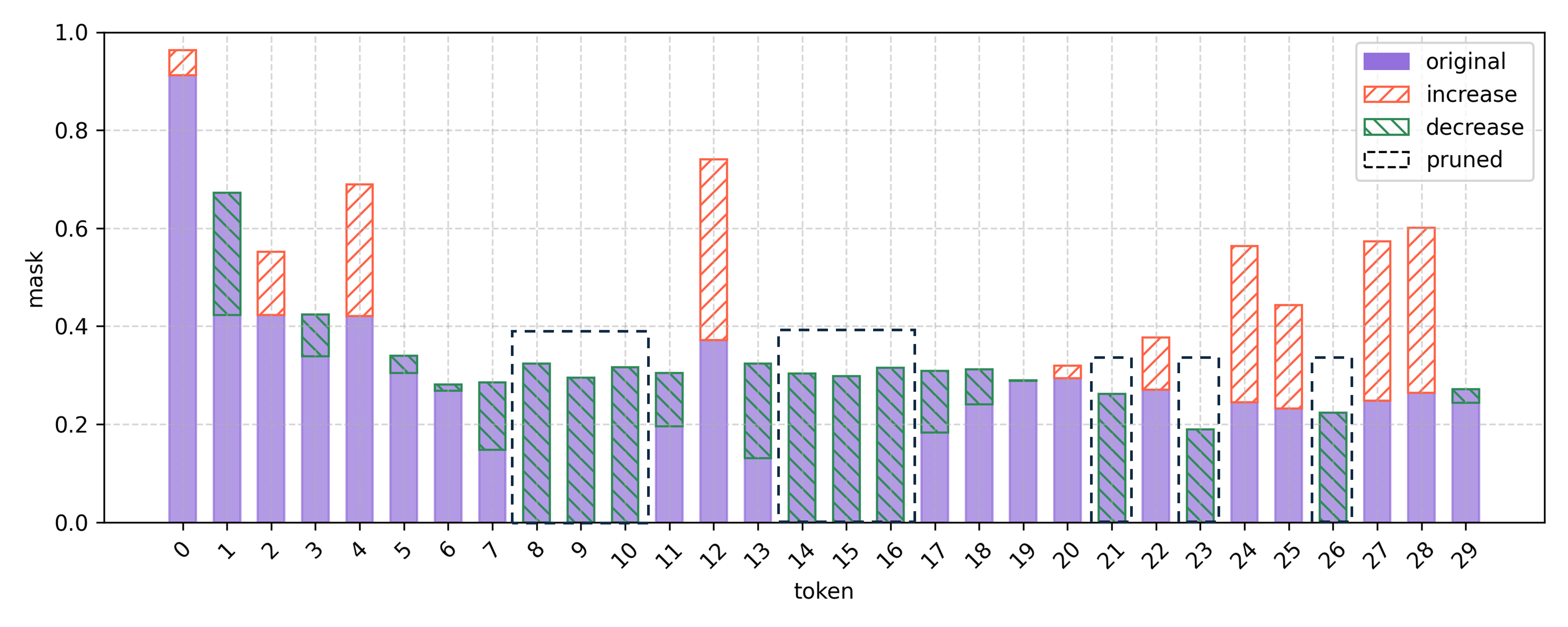}
\caption{Mask Evolution in Harmful Query Attacks. Initial and final mask values for each token position during attack on Llama-2-7B. Purple bars show initial values, red hatched areas indicate increased values, green areas show decreased values, and black dashed boxes highlight pruned tokens.}
\label{fig3}
\end{figure}

\subsection{Experimental Results}
We evaluate Mask-GCG by integrating it into three GCG variants across three LLMs, maintaining their original configurations. Tables~\ref{tab:RES1} and~\ref{tab:RES2} report the performance on SCR, ASR, and computational efficiency.

\noindent \textbf{SCR Analysis.} 
As shown in Table~\ref{tab:RES1}, Mask-GCG yields suffix compression across target models and GCG variants, with ratios up to $10.5\%$. The stability of this compression is supported by low standard deviations, indicating our method reliably identifies redundant tokens across different initializations. The degree of compression depends on the optimization characteristics of the underlying attack.

GCG+Mask-GCG achieves significant compression on Llama models, benefiting from extensive iterative pruning opportunities. While I-GCG's rapid convergence limits pruning windows on smaller models, substantial gains on Llama-2-13B suggest larger models expose more redundancy. AmpleGCG exhibits stable compression across architectures. Crucially, the impact of this pruning is non-linear in discrete optimization: even a small reduction in sequence length exponentially shrinks the search space, explaining how modest compression yields significant savings in gradient computation and sampling overhead.

\noindent \textbf{ASR Analysis.} 
Table~\ref{tab:RES2} presents ASR comparisons before and after integrating Mask-GCG across different target models and suffix lengths. Despite token pruning, Mask-GCG maintains or slightly improves ASR without compromising the effectiveness of original attack methods. Statistical analysis (95\% CI) confirms that observed fluctuations fall within the margin of error, indicating no significant performance loss.

Consistent ASR preservation patterns emerge across all experimental conditions. GCG+Mask-GCG maintains comparable attack success rates while achieving suffix compression. I-GCG+Mask-GCG exhibits near-perfect attack success rates matching or exceeding original I-GCG performance. AmpleGCG+Mask-GCG sustains the high performance levels characteristic of the original method. Failures are limited to edge cases where aggressive pruning on short suffixes disrupts the minimal semantic triggers required to bypass refusal.

The robustness of these results across different model architectures and parameter scales validates our core hypothesis: redundant tokens exist in adversarial suffixes, and removing low-impact tokens does not compromise attack effectiveness.

\noindent \textbf{Computational Efficiency Analysis.} 
Table~\ref{tab:speed_comparison} compares computational costs, showing that Mask-GCG consistently achieves significant efficiency gains with average time reductions of $16.5\%$ and $17.0\%$ for suffix lengths of 20 and 30, respectively.

Efficiency gains are most pronounced on Llama-2-7b, while Llama-2-13b results demonstrate scalability across model sizes. Notably, longer suffixes benefit more from optimization as the expanded search space facilitates more effective redundancy identification. Even fast-converging models like Vicuna-7b show consistent improvements.

These improvements stem from pruning low-impact tokens, which shrinks the gradient space and eliminates evaluation overhead. The consistent performance across architectures confirms the practical utility of redundancy-based optimization for adversarial attack generation.

\noindent \textbf{Distribution of Masks.} 
Figure~\ref{fig2} illustrates a representative case where Mask-GCG compresses a 30-token suffix to 18, achieving a $40\%$ compression ratio. The high probability values of retained tokens confirm the method's precision in selecting components exceeding the pruning threshold.

Further revealing the optimization dynamics, Figure~\ref{fig3} tracks the evolution of mask values. The clear divergence validates the adversarial suffix redundancy hypothesis by showing that critical positions gain weight while others decay, thereby demonstrating the effective separation of essential signals from noise.

\noindent \textbf{Stealthiness Analysis.} 
We conducted perplexity analysis on 600 adversarial suffixes across three models using GPT-2 \cite{GPT2} as an evaluator. Results show a median perplexity decrease from $4583.5$ (GCG) to $3486.4$ (Mask-GCG), achieving a $24\%$ reduction. This validates that pruning low-impact tokens improves the signal-to-noise ratio, producing more linguistically coherent suffixes. Crucially, while our experiments focus on open-weights models, this improved naturalness suggests Mask-GCG may generate suffixes that transfer better to black-box commercial models, which often employ perplexity-based filters for defense.

\section{Conclusion}
\label{sec:conclusion}
We investigate whether all tokens in adversarial suffixes are necessary for successful jailbreak attacks, revealing significant token redundancy within fixed-length suffixes. We propose Mask-GCG, a plug-and-play method using learnable token masking to identify impactful tokens in adversarial suffixes, addressing three key issues from low-impact tokens: attentional distraction, computational overhead, and stealthiness issues. Experimental results across models and variants demonstrate that Mask-GCG achieves a 16.8\% average time reduction while maintaining ASR. Specifically, we find that high-impact tokens account for approximately 83\% of the suffix, allowing for 7.5\% compression through selective pruning. Perplexity analysis on adversarial suffixes shows a 24\% reduction, confirming that removing redundant tokens improves the signal-to-noise ratio and enhances attack stealthiness. Through Mask-GCG, we demonstrate that learnable token masking effectively balances attack effectiveness with computational efficiency and stealthiness. The discovery of token redundancy opens new avenues for prompt optimization and provides deeper insights into how language models process adversarial inputs, contributing to both attack efficiency and defense mechanisms \cite{ai2023information, kong2025token}. For defense, the isolation of high-impact tokens suggests a shift from heuristic blocks to precise, importance-based filtering. Finally, the plug-and-play nature of our framework ensures applicability across diverse methods, establishing a solid foundation for future research in adversarial suffix optimization.

\newpage
\section{Ethical Considerations}
Our objective is to advance understanding of LLM vulnerabilities and contribute to stronger defensive mechanisms. While our results demonstrate efficiency improvements in adversarial attack generation, we emphasize that this work is intended solely for security research and defense development. We encourage researchers and practitioners to apply these findings toward building robust safeguards, including improved detection methods, model hardening, and comprehensive safety measures for ethical AI development.
\bibliographystyle{IEEEbib}
\bibliography{strings,refs}
\end{document}